\documentclass[sigconf,nonacm]{aamas}

\title{Constrained Black-Box Attacks Against Cooperative Multi-Agent Reinforcement Learning}

\author{Amine Andam}
\affiliation{
  \institution{Mohammed VI Polytechnic University}
  \country{Morocco}}
\email{andamamine83@gmail.com}

\author{Jamal Bentahar}
\affiliation{
  \institution{Khalifa University}
  \institution{Concordia University}
  \country{UAE}}
\email{jamal.bentahar@ku.ac.ae}
\author{Mustapha Hedabou}
\affiliation{
  \institution{Mohammed VI Polytechnic University}
  \country{Morocco}}
\email{mustapha.hedabou@um6p.ma}

\begin{abstract}
  Collaborative multi-agent reinforcement learning (MARL) is increasingly deployed in critical domains, yet its vulnerability to adversarial perturbations has not been thoroughly investigated. Most existing adversarial attacks against MARL rely on optimistic assumptions, such as access to policy weights or the ability to train surrogate policies. We investigate whether adversaries with limited capabilities remain effective against MARL agents. In particular, we study two restricted attack settings that are overlooked in the literature: (i) adversaries with access only to agent observations and (ii) adversaries with no access whatsoever. Our approach generates perturbations that intentionally misalign how victim agents perceive their environment, preventing them from collaborating. We conduct extensive experiments across a diverse set of 22 environments. Our results demonstrate both the effectiveness and sample efficiency of our attacks.
\end{abstract}


\newcommand{\BibTeX}{\rm B\kern-.05em{\sc i\kern-.025em b}\kern-.08em\TeX}

\begin{document}




\maketitle
\pagestyle{plain}
\thispagestyle{plain}

\section{Introduction}

Collaborative multi-agent reinforcement learning (c-MARL) algorithms have demonstrated state-of-the-art performance in complex cooperative tasks~\cite{rashid2018qmix,yu2022surprising}, making them well-suited for solving real-world problems across various domains \cite{DBLP:journals/tnn/LvLY25,10413998,9779871}. However, a critical prerequisite for widespread adoption of c-MARL is a full understanding of its vulnerabilities to adversarial attacks against policy networks \cite{huang2017adversarial,kos2017delving}, particularly when deployed.

We study test-time adversarial attacks against c-MARL agents, where an adversary manipulates the agents' observations by adding small perturbations. Existing test-time attacks have primarily been studied under white-box threat models \cite{pham2023attacking,lin2020robustness,nisioti2021robust}, which assume access to the policy architecture and parameters. In contrast, black-box threat models assume no access to either. They generally rely on learning a surrogate policy to exploit the transferability of adversarial perturbations, meaning that different policies trained for the same task are often fooled by the same adversarial perturbations. The surrogate policy can be learned either by training it from scratch \cite{huang2017adversarial} or by imitation learning \cite{wu2021adversarial,inkawhich2020snooping}. The former requires access to the training environment, whereas the latter requires collecting observation-action pairs.

In this work, we study adversarial attacks from a threat-modeling perspective. A threat model specifies the assumptions made about the attacker, particularly what information is available to generate the perturbations (e.g., weights, environment). In general, threat models that make weaker knowledge assumptions are more broadly applicable because they require less information about the target system. For example, assuming access only to observations is more broadly applicable than assuming access to observation-action pairs, which in turn is more plausible than assuming access to the policy weights. However, reducing the attacker's knowledge also makes attacks more difficult to design and typically less effective.

\begin{figure}[t]
    \centering
    \includegraphics[width=\columnwidth]{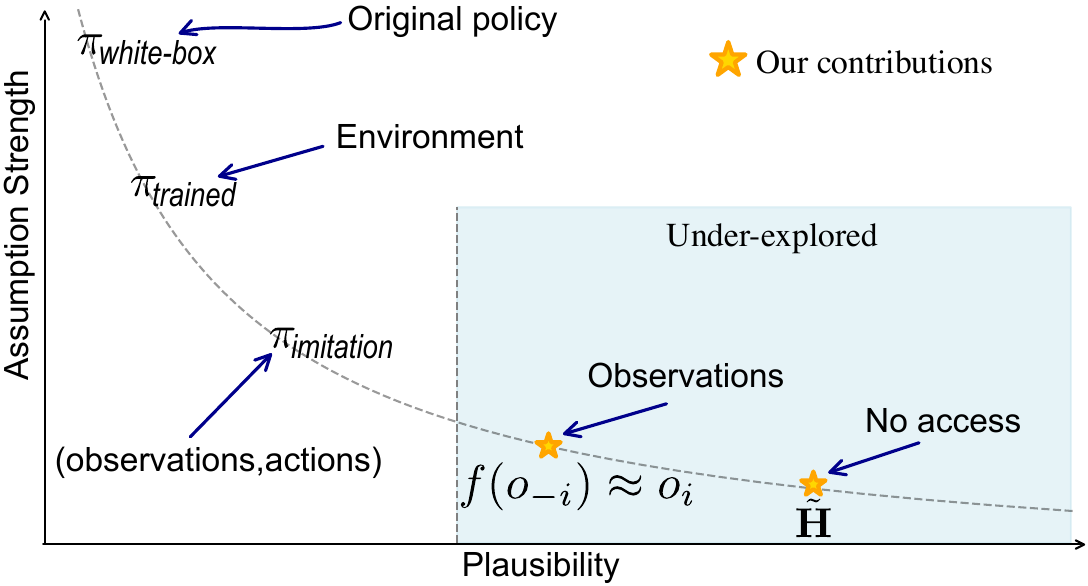}
    \caption{Attacker Knowledge vs Plausibility. }
    \Description{Attacker Knowledge vs Plausibility: There is an inverse relationship between attacker knowledge and threat-model plausibility: the less information an attacker is assumed to possess, the more plausible the corresponding scenario becomes. }
    \label{fig:realistic}
\end{figure}

Among the prior test-time attacks against c-MARL, the most constrained black-box threat models, excluding noise-based attacks, assume access to observation-action pairs. This is not a coincidence. Current attacks rely on adversarial transferability, which requires a surrogate policy, and learning such a policy requires at least observation-action pairs. As a result, threat models that assume access to less information than observation-action pairs have remained unexplored. Our research question is therefore: \emph{Can effective black-box attacks against c-MARL be generated when the available information is insufficient to learn a surrogate policy?}

The surrogate policy provides a signal to guide perturbation generation. For example, an attacker can reduce the probability of an agent's optimal action $a_i^{*}$ by solving:
\[
    \delta_i = \operatorname*{arg\,min}_{\|\delta_i\|_{\infty} \leq \epsilon} \tilde{\pi}(a_i^{*} \mid o_i + \delta_i)
\]
where $\tilde{\pi}$ denotes the surrogate policy. The absence of this policy leaves the adversary without guidance on how to generate harmful perturbations. Addressing our research question therefore requires an alternative attack objective that can be constructed from the limited information the adversary can access.

Our first contribution is to show that an adversary with access only to agents' observations (and not their actions) can effectively attack c-MARL agents. 
Since we target cooperative agents, we exploit the fact that receiving consistent information helps them maintain aligned beliefs about their environment, on which their ability to coordinate depends. Our goal is therefore to reduce the amount of consistent information available to the agents through small perturbations that misalign their beliefs and prevent them from coordinating. For example, two agents that must jointly intercept a moving target may fail to coordinate if the perturbations make each of them perceive the target at a different location. To construct the attack objective, we draw on the intuition that aligned observations tend to be correlated as they encode similar information. We learn these correlations by training a neural network to predict each agent's observation from those of the remaining agents. We use the reconstruction error as a proxy for misalignment and search for perturbations that maximize this error. We refer to this method as the Misalign attack.

Our second contribution is an attack under an even more constrained threat model, in which the adversary can only inject perturbations into the agents’ observations, without being able to read or collect them. Such attacks are typically implemented by injecting random noise. Our new attack, which we call the Hadamard attack, uses partial Hadamard matrices to generate the perturbations. As we discuss later, these matrices produce orthogonal perturbations, which we use as a heuristic motivated by the same misalignment intuition as our first attack.

These two attacks can also be combined by using observations to select a subset of agents and partial Hadamard matrices to efficiently generate perturbations. An overview of the proposed attacks is shown in Figure~\ref{fig:realistic}.

Finally, we extensively evaluate our attacks across three MARL benchmark suites spanning 22 tasks, including fully observable, partially observable, and highly cooperative environments. Our results show that effective attacks remain possible without the information required to train surrogate policies. In several tasks, Misalign matches or outperforms imitation learning based attacks despite requiring less information. We also conduct a comprehensive ablation study covering policy-based and value-based victim algorithms, network architectures, and data collection requirements. In particular, we show that observations collected over as few as 1,000 environment steps can suffice to generate effective Misalign attacks.

\section{Background and Related Work}

\subsection{Background}
Consider a neural network $f$ parameterized by $\theta$, trained to minimize a loss function $J(x,y;\theta)$, where $x \in \mathcal{X}$ is the input and $y$ is its corresponding label. An adversarial attack aims to add a perturbation $\delta$ to the input $x$ that maximizes the loss $J(x + \delta, y; \theta)$. To avoid detection, $\delta$ must remain small, which is enforced by constraining its $L_{\infty}$ norm with a budget $\epsilon$. This problem is formulated as:
\begin{equation}
    \delta = \arg\max_{\delta} J(x + \delta, y; \theta) \quad \text{subject to} \quad \|\delta\|_{\infty} \leq \epsilon, \quad x+\delta \in \mathcal{X}
    \label{eq:background}
\end{equation}

A widely used method for solving this problem is Projected Gradient Descent (PGD) \cite{kurakin2018adversarial,mkadry2017towards}. PGD applies iterative perturbations over $K$ steps with a step size $\alpha$:  
\begin{equation}
    x^{*}_{0} = x, \quad x^{*}_{t+1} = Clip_{\epsilon, \mathcal{X}} \left\{ x^{*}_{t} + \alpha \cdot \mathrm{sign}\!\left(\nabla_{x} J(x^{*}_{t}, y; \theta)\right) \right\}
    \label{eq:pgd}
\end{equation}
\noindent where $\nabla_{x} J$ is the gradient of the loss with respect to the input, $\mathrm{sign}(\cdot)$ returns $+1$ for a positive argument, $-1$ for a negative argument, and $0$ otherwise, and $Clip_{\epsilon, \mathcal{X}}$ is an element-wise clipping operator that keeps the perturbation within the $\epsilon$ budget and the input within the valid domain $\mathcal{X}$.

\subsection{Related Work}
\label{sec:relatedwork}
\begin{table*}[th]
    \centering 
    \caption{Comparison of our work and previous papers.}
    \setlength{\tabcolsep}{1mm}
    \begin{tabular}{lcc ccccc cccc}
        \textbf{Paper}  & \textbf{Test-time} & \textbf{Black-box} & \multicolumn{5}{c}{\textbf{Attacker Knowledge}} & \multicolumn{4}{c}{\textbf{Target}} \\
        \midrule
        & & & Policy & Obs & Actions & Reward & Env & Actions & Obs & Reward & Env \\
        \cmidrule(lr){4-8} \cmidrule(lr){9-12} 
        Pham et al. (2023)  & \ding{51} & \ding{55} & \ding{51} & \ding{51} & \ding{51}  & \ding{51} & \ding{51} & \ding{55} &\ding{51}  & \ding{55}  & \ding{55}   \\
        Lin et al. (2020) & \ding{51} & \ding{55} & \ding{51} & \ding{51} & \ding{51}  & \ding{51} & \ding{51} & \ding{55} &\ding{51}  & \ding{55}  & \ding{55}   \\
        Nisioti et al. (2021)  & \ding{51} & \ding{55} & \ding{51} & \ding{51} & \ding{51}  & \ding{51} & \ding{55} & \ding{51} &\ding{55}  & \ding{55}  & \ding{55}  \\
        \citeauthor{DBLP:conf/aaai/ChenWWXHWX26}~(\citeyear{DBLP:conf/aaai/ChenWWXHWX26})& \ding{51} & \ding{51} & \ding{55} & \ding{51} & \ding{51}  & \ding{51} & \ding{51} & \ding{55} &\ding{51}  & \ding{55}  & \ding{55}  \\
        Hu and Zhang (2022)  & \ding{55} & \ding{55} & \ding{51} & \ding{51} & \ding{51}  & \ding{51} & \ding{51} & \ding{51} &\ding{55}  & \ding{55}  & \ding{55}   \\
        Chen et al. (2024) & \ding{55} & \ding{55} & \ding{51} & \ding{51} & \ding{51}  & \ding{51} & \ding{51} & \ding{51} &\ding{55}  & \ding{55}  & \ding{55}   \\
        Liu and Lai (2023)  & \ding{55} & \ding{51} & \ding{55} & \ding{51} & \ding{51}  & \ding{51} & \ding{51} & \ding{51} &\ding{55}  & \ding{51}  & \ding{55}  \\
        Zheng et al. (2023)  & \ding{55} & \ding{51} & \ding{55} & \ding{51} & \ding{51}  & \ding{51} & \ding{55} & \ding{55} &\ding{51}  & \ding{55}  & \ding{55}  \\
        \textcolor{red}{\textbf{Ours}} & \textcolor{red}{\ding{51}} & \textcolor{red}{\ding{51}} & \textcolor{red}{\ding{55}} & \textcolor{red}{(\ding{51},\ding{55})} & \textcolor{red}{\ding{55}}  & \textcolor{red}{\ding{55}} & \textcolor{red}{\ding{55}} & \textcolor{red}{\ding{55}} & \textcolor{red}{\ding{51}}  & \textcolor{red}{\ding{55}}  & \textcolor{red}{\ding{55}}  \\
        \midrule
        \textcolor{red}{}
    \end{tabular}
    
    \label{tab:comp}
\end{table*}
We organize prior work on adversarial c-MARL along two axes: whether the attack occurs at training or test time, and whether it assumes a white-box or black-box threat model.

\textbf{\textit{Training-time vs Test-time attacks}}. Training-time attacks, also called data poisoning attacks \cite{rakhsha2020policy}, occur when an adversary is present during training and aim to manipulate the agent into learning a target policy chosen by the adversary instead of the optimal policy. This can involve reward poisoning \cite{liu2023efficient,zhang2020adaptive}, environment poisoning \cite{rakhsha2021policy}, or altering observations or actions \cite{hu2022sparse,chen2024cuda2,zheng2023one4all}. However, interfering with training is not always feasible. MARL systems are arguably more vulnerable when deployed. Test-time attacks occur during deployment and aim to degrade agent performance. This is achieved by exploiting the known vulnerabilities of neural networks to adversarial inputs \cite{huang2017adversarial,szegedy2013intriguing}, mainly through observation manipulation \cite{pham2023attacking,lin2020robustness}, or action manipulation  \cite{nisioti2021robust}.

\textbf{\textit{White-box vs Black-box}}. In white-box scenarios, the attacker knows the learning algorithm and has access to the policy’s weights and architecture \cite{chen2024cuda2,hu2022sparse,nisioti2021robust,lin2020robustness,pham2023attacking,huang2017adversarial}. While these attacks tend to be the most effective, it is impractical for the attacker to have complete knowledge of the deployed policies. Conversely, black-box settings allow for more relaxed assumptions \cite{huang2017adversarial}. Most existing work relies on learning a surrogate policy to exploit the transferability of adversarial examples \cite{papernot2016transferability}. This surrogate policy can be learned by training the model from scratch \cite{huang2017adversarial}, but doing so requires access to the original training environment. Alternatively, imitation learning can be used to approximate the policy \cite{DBLP:conf/aaai/ChenWWXHWX26,wu2021adversarial,inkawhich2020snooping}, which would necessitate access to observation-action pairs or the ability to query the policy.

Table~\ref{tab:comp} compares our work with prior attacks. Existing test-time attacks typically assume broad attacker knowledge, which is often unavailable at deployment. We instead target a more constrained black-box setting with limited knowledge.

\section{Methodology}
\subsection{ Threat Model}

We consider a c-MARL system at test time. Let $\mathcal{N}$ be a set of $n$ agents. We denote by $o_i \in \mathbb{R}^d $ the observation of agent $i$, by $o_{-i}$ the concatenation of all observations except that of agent $i$, and by $\mathbf{o}$ the joint observation. As in standard observation-perturbation attacks, we assume that the adversary can inject additive perturbations into the agents' observations. The adversary knows the number of agents, the observation dimensions, and the numerical bounds of the observation space. Our focus is on how to generate effective perturbations given this ability, rather than on how the adversary obtains it.

We consider the following two threat models (TMs):
\begin{itemize}
    \item \textbf{Threat model 1 (TM1)}: The adversary can \emph{collect and perturb} the agents' observations.
    \item \textbf{Threat model 2 (TM2)}: The adversary can \emph{only perturb} the agents' observations, without being able to read or collect them.
\end{itemize}

In neither threat model does the adversary have access to the agents' actions, rewards, policy architecture or parameters, policy queries, or the training environment. We constrain the perturbations using the $L_{\infty}$ norm and an attack budget $\epsilon$.

The adversary's capabilities in our two threat models have close counterparts in cybersecurity. TM1 is analogous to man-in-the-middle attacks, where an adversary can observe transmitted information before manipulating it. TM2 is analogous to sensor-spoofing attacks (e.g., GPS or LiDAR spoofing), where an adversary can modify sensor readings without observing the original measurements.

\subsection{Intuition}
\begin{figure}[t]
    \centering
    \includegraphics[width=\columnwidth]{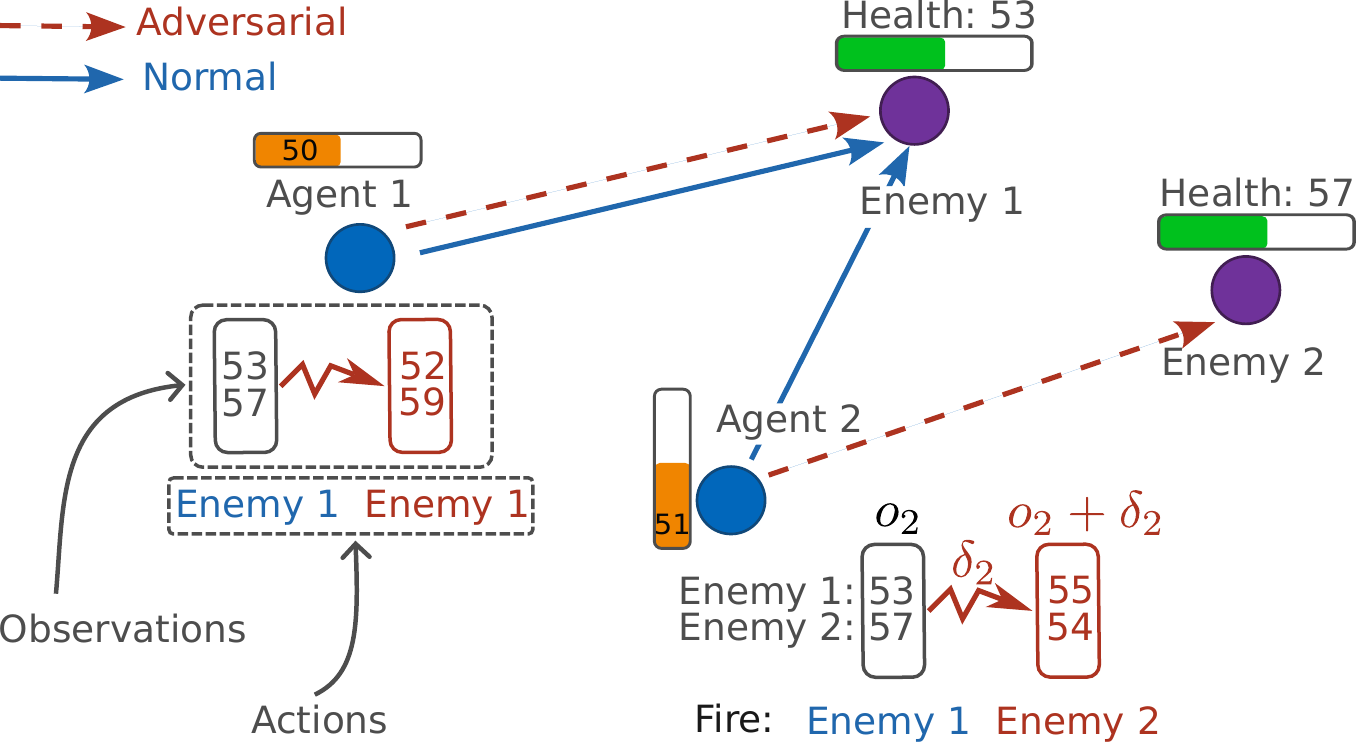}
    \caption{Perturbations make different agents perceive different enemies as having the lowest health. }
    \Description{Perturbations make different agents perceive different enemies as having the lowest health.}
    \label{fig:illustration}
\end{figure}

Successful cooperation requires agents to make coordinated decisions, which in turn depends on them having a consistent understanding of the environment. For example, consider the SMAC benchmark \cite{DBLP:conf/atal/SamvelyanRWFNRH19,ellis2024smacv2}, where agents often cooperate through \textit{focus fire}: they jointly attack the weakest enemy to eliminate it quickly. This requires the agents to agree on the target's health and location. As illustrated in Figure~\ref{fig:illustration}, if adversarial perturbations cause different agents to perceive different enemies as the weakest, they may attack different targets instead of coordinating.

Our attack exploits this dependency by trying to create disagreement between the agents. Our idea is similar to a divide-and-conquer strategy: we aim to disrupt cooperation by causing the agents to perceive the environment differently, making it harder for them to coordinate their decisions.

This intuition also extends to partially observable environments, where there may be little or no overlap between agents' observations. In such settings, coordination relies on agents having aligned common beliefs about the environment \cite{DBLP:journals/ijon/ZhangLMY22,DBLP:journals/corr/abs-2102-02274}. We focus our experiments on these partially observable scenarios, as they are more challenging for our attacks.

\subsection{Threat Model 1: Misalign Attack}
\label{sec:align}

Since our goal is to transform aligned observations into misaligned ones, we first need a quantitative measure of how aligned the observations are. Our key insight is that aligned observations tend to be correlated, because they encode similar information about the environment. As a result, the observation of one agent can be approximated from the observations of the remaining agents.

We learn this mapping with a neural network $f_{\theta}: \mathbb{R}^{(n-1)d} \to \mathbb{R}^{d}$, which we call the \emph{alignment network}. It is trained to predict $o_i$ from $o_{-i}$ by minimizing the reconstruction error:

\begin{equation}
    \mathcal{L}(\mathbf{o}; \theta) = \frac{1}{n} \sum_{i \in \mathcal{N}} \left\| f_{\theta}(o_{-i}) - o_{i} \right\|_2^2
    \label{eq:align_loss}
\end{equation}

Just as aligned observations tend to be correlated and mutually predictable (a low reconstruction loss), misaligned observations will be less correlated and harder to predict (a higher reconstruction loss). Therefore, we aim to transform aligned observations into misaligned ones by disrupting these correlations, using a higher reconstruction loss as a proxy for misalignment: 
\begin{equation}
    \mathcal{L}(\mathbf{o}; \theta) \leq \mathcal{L}(\mathbf{o} + \boldsymbol{\delta}; \theta)
\end{equation}

Accordingly, we search for the perturbation that maximizes this loss. Our attack is formulated as the following optimization problem:
\begin{equation}
    \boldsymbol{\delta} = \arg\max_{\boldsymbol{\delta}}\; \mathcal{L}(\mathbf{o} + \boldsymbol{\delta}; \theta) \quad \text{s.t.} \quad \|\boldsymbol{\delta}\|_{\infty} \leq \epsilon
    \label{eq:method}
\end{equation}
\noindent where
\begin{equation}
    \mathcal{L}(\mathbf{o} + \boldsymbol{\delta}; \theta) = \frac{1}{n} \sum_{i \in \mathcal{N}} \left\| f_{\theta}(o_{-i} + \delta_{-i}) - (o_{i} + \delta_i) \right\|_2^2
    \label{eq:method_loss}
\end{equation}

The problem in Eq.~(\ref{eq:method}) can be efficiently optimized using PGD. We refer to this method as the \emph{Misalign attack}. It consists of two phases: a data collection phase, in which the attacker collects observations to train the alignment network, and an attack phase, in which the network parameters are frozen and adversarial perturbations are generated at each time step.

Sometimes, it is desirable to attack only a subset of agents $\mathcal{M} \subset \mathcal{N}$, for example to reduce the risk of detection. In this case, we use the reconstruction loss to select which agents to target. For a fixed number of targeted agents $m$, we select the $m$ agents with the lowest loss values $\left\| f_{\theta}(o_{-i}) - o_{i} \right\|_2^2$. We interpret them as the agents whose observations contain information upon which most agents agree. Consequently, perturbing these agents is expected to reduce the overall alignment. Formally, we select the subset $\mathcal{M}$ as follows:
\begin{equation}
    \mathcal{M} =
    \operatorname*{arg\,min}_{\substack{\mathcal{S} \subset \mathcal{N} \\ |\mathcal{S}| = m}}
    \sum_{i \in \mathcal{S}} \left\| f_{\theta}(o_{-i}) - o_{i} \right\|_2^2
    \label{eq:alignselection}
\end{equation}
 
Overall, our attack bypasses the requirement of training a surrogate policy to generate black-box perturbations. Instead, we train an alternative network using only observations. 
\subsection{Threat Model 2: Hadamard Attack}
\label{sec:hadamard}

Our goal is to break the alignment of the observations without being able to read them. To illustrate our method, imagine a two-dimensional game in which two agents must jointly catch an object. Each agent observes the $(x, y)$ coordinates of the target. To prevent them from completing their task, it is enough to make them pursue completely dissimilar directions. A suitable strategy is to push the agents in perpendicular directions. We did not choose opposite directions because this idea cannot be generalized to more than two agents. We cannot have a set of three or more vectors in which every pair points in opposite directions.

We propose a heuristic attack that generalizes this idea to high-dimensional observation spaces using \emph{vector orthogonality}. For $n$ agents with observations of dimension $d$, we generate a perturbation matrix $\boldsymbol{\delta} \in \mathbb{R}^{n \times d}$ whose $i$-th row $\delta_i$ is added to agent $i$'s observation. We require the perturbations to satisfy two conditions:
\begin{align}
    \textit{(C1) Orthogonality:} \quad & \delta_i^\top \delta_j = 0
        \quad \forall i \neq j \in \mathcal{N} \label{cd1} \\
    \textit{(C2) Budget:} \quad & \|\delta_i\|_{\infty} \leq \epsilon
        \quad \forall i \in \mathcal{N} \label{cd2}
\end{align}

Generating such a matrix is not trivial. A greedy approach that constructs the rows one at a time easily meets the budget, but enforcing orthogonality against all previous rows becomes difficult in high-dimensional spaces. Alternatively, orthogonalizing a random matrix by QR decomposition guarantees \textit{(C1)}, but the resulting rows may have small magnitudes and leave most of the budget unused.

To generate perturbations that satisfy both conditions, we use partial Hadamard matrices. A Hadamard matrix $\mathbf{H} \in \mathbb{R}^{d \times d}$ has entries in $\{+1,-1\}$ and satisfies $\mathbf{H}\mathbf{H}^\top = d\,\mathbf{I}$, so its rows are orthogonal. Keeping any $n \leq d$ of them yields a partial Hadamard matrix $\tilde{\mathbf{H}} \in \mathbb{R}^{n \times d}$ that preserves this property. Scaling it by the attack budget gives rows that are pairwise orthogonal, satisfying \textit{(C1)}, and whose entries all have magnitude exactly $\epsilon$, attaining the maximum allowed $L_\infty$ norm in \textit{(C2)}:
\begin{equation}
    \label{eq:scaled_haddd}
    \boldsymbol{\delta} = \epsilon \, \tilde{\mathbf{H}}
\end{equation}

We refer to this attack as the \emph{Hadamard attack}. We generate Hadamard matrices using \textit{Sylvester’s construction}. Hadamard matrices exist only if $d$ is equal to 2 or a multiple of 4. To bypass this, we first generate a full Hadamard matrix of size $\tilde{d}$, where $\tilde{d}$ is the largest power of two such that $\tilde{d} \leq d$: 
\begin{equation}
    \tilde{d} = 2^{\left\lfloor \log_2 d \right\rfloor}
\end{equation}
and we pad the remaining $d - \tilde{d}$ columns with zeros. The resulting perturbations remain pairwise orthogonal and satisfy the attack budget. Every nonzero entry has magnitude $\epsilon$, while the padded entries are zero.

\section{Experiments}
\begin{figure*}[t]
    \centering
    \includegraphics[width=0.9\textwidth]{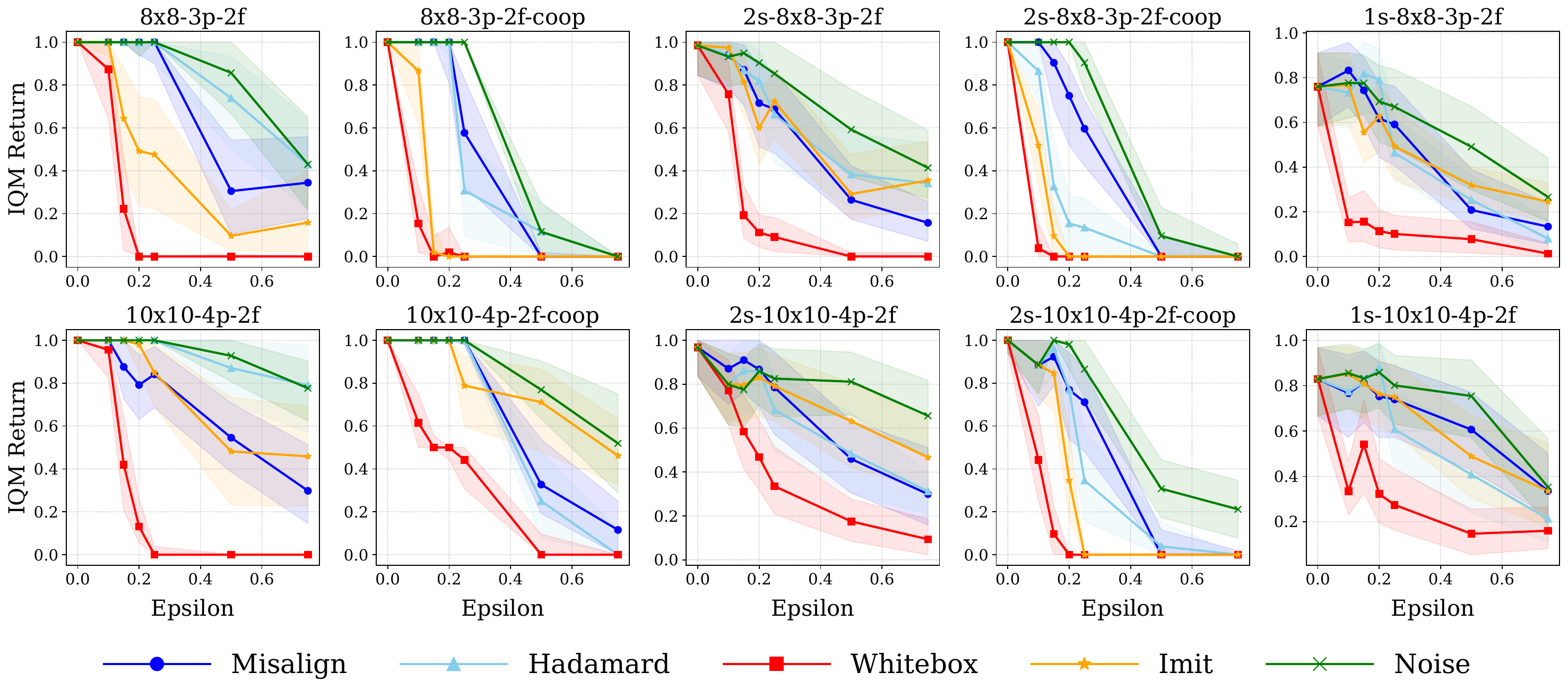}
    \caption{Performance on LBF.}
    \Description{Performance on LBF: IQM returns and 95\% CIs}
    \label{fig:ben_lbf}
\end{figure*}

The goal of this section is to empirically answer our research question: can an adversary effectively attack c-MARL without sufficient information to learn a surrogate policy? We evaluate the proposed attacks under both threat models across different tasks. We further investigate attacks on subsets of agents and examine how performance depends on the victim algorithm, data collection requirements, and alignment-network architecture.

\subsection{Experimental Setup}
\label{sec:exp_setup}
\subsubsection{Environments}

We carefully choose multiple environments from three different MARL benchmarks:

\begin{itemize}
    \item Level-Based Foraging (LBF)~\cite{DBLP:conf/nips/PapoudakisC0A21}:  a grid-world where agents collect randomly scattered food. Agents either see the whole grid or only a limited area around themselves: a 2-square (``-2s'') or 1-square (``-1s'') radius. In the highly cooperative setting (``-coop''), food can only be collected if all agents act on it simultaneously; coordination therefore depends critically on agents receiving identical information.
    \item Multi-Robot Warehouse (RWARE)~\cite{DBLP:conf/nips/PapoudakisC0A21}: a partially observable environment where robots collect requested shelves in a warehouse. Agents observe a $7\times7$ (``-sr-3''), $5\times5$ (``-sr-2''), or $3\times3$ (``-sr-1'') grid around them. Compared to LBF, RWARE has higher-dimensional observations.
    \item StarCraft Multi-Agent Challenge (SMAC and SMACv2)~\cite{DBLP:conf/atal/SamvelyanRWFNRH19,ellis2024smacv2}: provides partially observable combat scenarios with a large number of agents, which lets us evaluate how our attacks scale. Since agents may die during combat, the attacker progressively loses access to agents. We evaluate three scenarios from SMAC~\cite{DBLP:conf/atal/SamvelyanRWFNRH19}: \texttt{10m\_vs\_11m}, \texttt{27m\_vs\_30m}, and \texttt{MMM2}; and three from SMACv2~\cite{ellis2024smacv2}: \texttt{protoss\_5\_vs\_5}, \texttt{terran\_5\_vs\_5}, and \texttt{zerg\_5\_vs\_5}.
\end{itemize}

We train the agents using MAPPO~\cite{yu2022surprising} and QMIX~\cite{rashid2018qmix}. We evaluate three architectures for $f_\theta$: MLP, RNN, and encoder-only Transformers. Unless otherwise stated, we report results against MAPPO agents using an RNN alignment network trained on observations collected over 5,000 environment steps. We use $K=10$ PGD iterations with step size $\alpha=\frac{\epsilon}{K}$. Additional details about the environments and hyperparameters are in the Appendix.

\subsubsection{Adversarial Baselines}
We use the following baselines:
\begin{itemize}
    \item \textit{White-box attack:} The adversary has full access to the policy. We generate perturbations that minimize the probability (or the Q-value for QMIX) of the optimal action:
    \[
    \delta_i = \operatorname*{arg\,min}_{\|\delta_i\|_{\infty} \leq \epsilon} \pi(a_i^{*} \mid o_i + \delta_i)
\]
    \item \textit{Imitation attack:}  This black-box attack trains a surrogate policy $\pi^{\mathrm{imit}} $. We use 5,000 environment steps and an RNN similar to the one used to train $f_{\theta}$. 
    \[
    \delta_i = \operatorname*{arg\,min}_{\|\delta_i\|_{\infty} \leq \epsilon} \pi^{\mathrm{imit}}(a_i^{*} \mid o_i + \delta_i)
\]
    \item \textit{Random attacks}: Most previous works used only uniform distributions $\delta_i \sim \mathcal{U}(-\epsilon, \epsilon)$ and normal distributions $\delta_i \sim \mathcal{N}(0, \epsilon^2)$. We add Rademacher perturbations in which each entry is independently sampled from $\{-\epsilon,+\epsilon\}$ with equal probability. This provides a random-sign baseline whose entries all have magnitude $\epsilon$. Hadamard perturbations have the same magnitude on their nonzero entries, while the dimensions added by zero-padding remain unperturbed. In addition, we add a non-symmetric distribution and a temporally correlated noise: for the former, we use the exponential distribution $\mathrm{Exp}(\lambda)$. The value of $\lambda$ is chosen such that 99\% of the sampled values satisfy the budget constraint:
    \begin{equation}
        \lambda = - \frac{\ln(0.01)}{\epsilon}
    \end{equation}
    For correlated noise, we use the Ornstein–Uhlenbeck (OU) process:
    \begin{equation}
        \delta_{t+1} = \delta_t + \theta (\mu - \delta_t) \Delta t + \sigma \sqrt{\Delta t} \, \mathcal{N}(0, 1)
        \end{equation} 
In our experiments, we use the following parameters:
\begin{equation*}
    \mu = \epsilon, \theta \approx \epsilon, \sigma \approx  \frac{\epsilon}{10} \pm 0.001,  \Delta t =1  
\end{equation*}
\end{itemize}

When presenting numerical results, and in order to avoid overloading the figures with curves of random attacks, we report for each $\epsilon$ the performance of the best random attack (i.e., the random noise with the lowest return), and refer to it as the "Random attack". For all attacks and baselines, perturbations are clipped element-wise to \([-\epsilon,\epsilon]\), and perturbed observations are clipped to the numerical bounds of the observation space

It is important to note that both the white-box and the imitation attack assume more knowledge than our attacks. To the best of our knowledge, there are no other attacks that have similar threat models to ours.

Finally, we report the Interquartile Mean (IQM)~\cite{DBLP:conf/nips/AgarwalSCCB21} and 95\% confidence intervals of episodic returns using 50 episodes. We use IQM because it is robust to outliers and less biased than the median. Our code is available at \url{https://github.com/AmineAndam04/black-box-marl}.

\subsection{Comparison with Baselines}
\label{sec:num_res}

\begin{figure*}[t]
    \centering
    \includegraphics[width=0.9\textwidth]{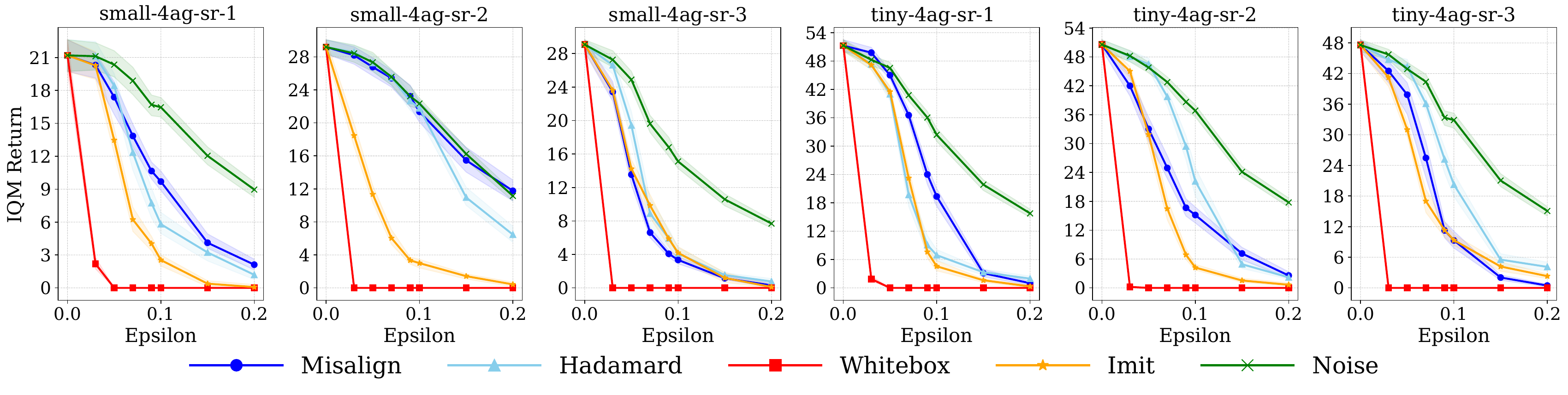}
    \caption{Performance on RWARE.}
    \Description{Performance on RWARE: IQM returns and 95\% CIs.}
    \label{fig:ben_rware}
\end{figure*}
\begin{figure*}[t]
    \centering
    \includegraphics[width=0.9\textwidth]{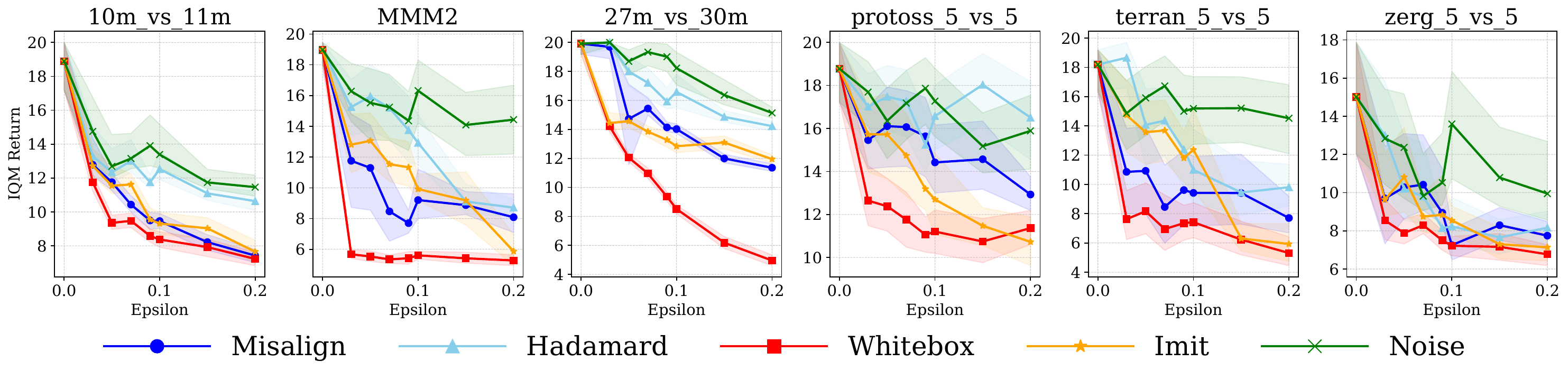}
    \caption{Performance on StarCraft. }
    \Description{Performance on StarCraft: IQM returns and 95\% CIs. }
    \label{fig:ben_smac}
\end{figure*}

Figures~\ref{fig:ben_lbf},~\ref{fig:ben_rware}, and~\ref{fig:ben_smac} present the IQM returns (y-axis) for each benchmark at different budgets $\epsilon$ (x-axis). The values at $\epsilon =0$ represent the benign performance.

\paragraph*{LBF} In fully observable scenarios (the first two columns in Figure~\ref{fig:ben_lbf}), our attacks significantly impact the agents' performance, particularly in highly cooperative tasks where the returns drop to very low values more rapidly. 
The same trend also holds under partial observability (compare 3rd and 4th columns). The Hadamard attack matches or outperforms the Misalign attack in partially observable scenarios (the last three columns), which we attribute to the difficulty of learning an accurate alignment network when observation correlations become weak. In 6 out of 10 tasks, one of our attacks matches or outperforms the Imitation attack, despite the latter assuming more knowledge.

\paragraph*{RWARE} In general, our attacks perform well and remain effective in the higher-dimensional observation spaces of this benchmark. Under the highest level of partial observability (1st and 4th columns in Figure~\ref{fig:ben_rware}), while the Misalign attack still performs well, it is slightly outperformed by the Hadamard attack. As the observation range increases (e.g., from 1st to 3rd and from 4th to 6th columns), Misalign becomes more effective, as $f_\theta$ becomes easier to learn and more useful. Except for \texttt{small-4ag-sr-2}, our attacks achieve performance close to that of the Imitation attack. As in the LBF tasks, the Hadamard attack consistently outperforms the Random attack (both have the same threat model).

\paragraph*{StarCraft} The Misalign attack maintains consistent performance across the StarCraft games. This confirms that our approach scales to cases with a high number of agents (the first three columns in Figure~\ref{fig:ben_smac}) and remains effective in situations with no persistent access to the agents. Except on \texttt{protoss\_5\_vs\_5}, it outperforms the Imitation attack at many attack budgets, and on \texttt{10m\_vs\_11m} and \texttt{zerg\_5\_vs\_5} it comes very close to the white-box attack. In contrast, the Hadamard attack struggles to significantly outperform the Random attack in many games: \texttt{10m\_vs\_11m}, \texttt{27m\_vs\_30m}, and \texttt{protoss\_5\_vs\_5}.

Overall, these results show that (1) effective black-box attacks remain possible even without the information needed to train a surrogate policy. Using only observations, Misalign often matches or outperforms imitation attacks and, in two StarCraft environments, approaches white-box performance. (2) Despite using no victim-specific information, the Hadamard attack is surprisingly effective, frequently matching imitation attacks in LBF and RWARE and outperforming the random baseline in many tasks, making it a stronger baseline for future work.

\subsection{Targeted Attacks}

\begin{figure}[thbp]
    \centering
    \includegraphics[width=\columnwidth]{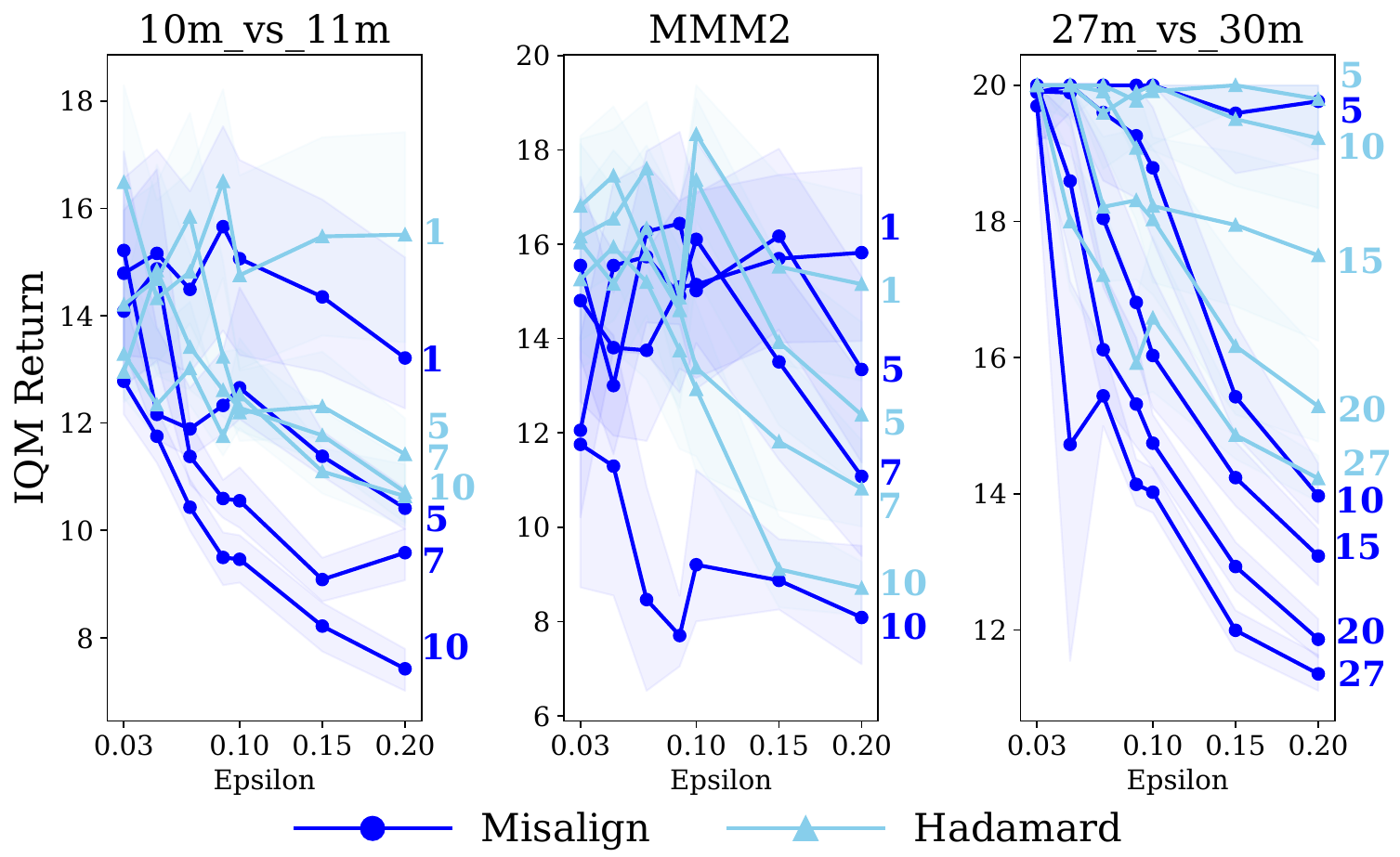}
    \caption{Performance of targeted attacks on StarCraft.}
    \Description{Performance of targeted attacks on StarCraft.}
    \label{fig:ben_smac_light}
\end{figure}

\begin{figure}[thbp]
    \centering
    \includegraphics[width=0.7\columnwidth]{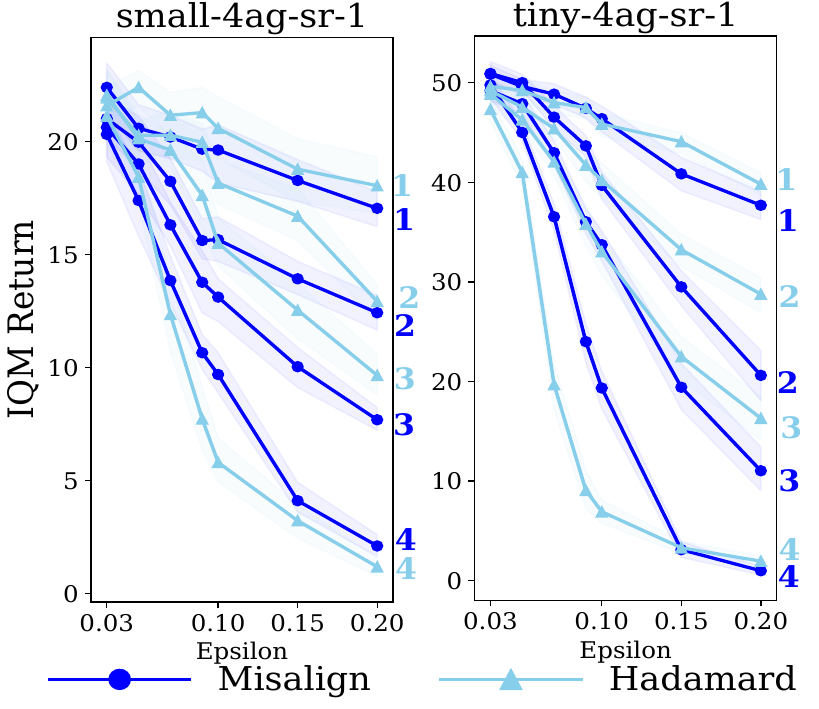}
    \caption{Performance of targeted attacks on RWARE. }
    \Description{Performance of targeted attacks on RWARE. }
    \label{fig:ben_rware_light}
\end{figure}

\begin{figure}[thbp]
    \centering
    \includegraphics[width=\columnwidth]{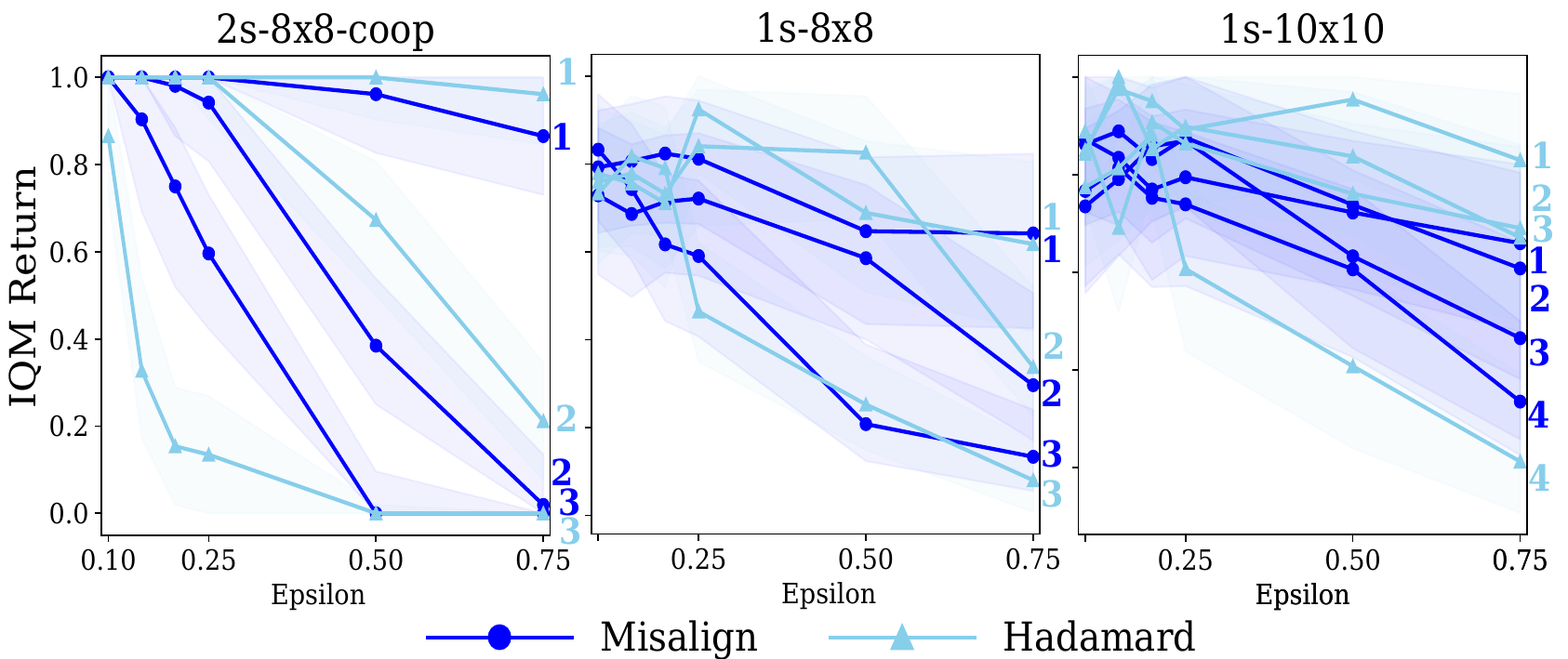}
    \caption{Performance of targeted attacks on LBF.}
    \Description{Performance of targeted attacks on LBF.}
    \label{fig:ben_lbf_light}
\end{figure}

We evaluate scenarios where we only attack a subset of selected agents. For the Misalign attack, we use the selection mechanism defined in Eq.~(\ref{eq:alignselection}), while for Hadamard we select agents randomly. Results are shown in Figures~\ref{fig:ben_smac_light},~\ref{fig:ben_rware_light}, and~\ref{fig:ben_lbf_light}. The number of attacked agents  is indicated on the right of each subplot. We do not include all the environments due to page limits, but we report the full results in the Appendix.

Two observations stand out: First, Misalign maintains strong performance when at least 50\% of the agents are targeted. Second, the Misalign attack is less affected by the reduction in the number of targeted agents than the Hadamard attack. To illustrate this, in Figures~\ref{fig:ben_rware_light} and~\ref{fig:ben_lbf_light} we show tasks where Hadamard outperforms Misalign when all agents are attacked, and we can see that Misalign becomes more effective than Hadamard as fewer agents are targeted.

These results suggest that the alignment network provides an effective mechanism for selecting agents to target. To further test this, we isolate the targeting mechanism and combine both attacks: we first select the target agents using Eq.~(\ref{eq:alignselection}) and then inject Hadamard perturbations instead of Misalign perturbations. We refer to this as targeted Hadamard. The results of the combined attacks are shown in Tables~\ref{tab:ben_lbf_targh} and ~\ref{tab:ben_rware_targh}. We report the additional drop in IQM returns when adding the selection mechanism to Hadamard: for example, if Hadamard alone causes a -15\% drop relative to the benign return and the value reported in the table is -10\%, then targeted Hadamard causes a -25\% drop relative to the benign return. In most of the benchmarks, the targeted Hadamard significantly outperforms the untargeted version, achieving a maximum additional drop of -57\%. Averaged over all evaluated tasks, including those reported in the Appendix, the additional drop is -11.5\% on the LBF benchmark, -6.18\% on RWARE tasks, and -6.28\% on StarCraft games. These results confirm the effectiveness of the alignment network for target selection.

\begin{table}[thbp]
    \centering
    \caption{Targeted Hadamard attack on LBF. }
    \begin{tabular}{lc ccc}
    \toprule
    Task & $m$ & $\epsilon = 0.15$ & $\epsilon = 0.2$ & $\epsilon = 0.25$  \\
    \midrule
    2s-8x8-coop   & 1 & 0 & -3.84 & -3.84  \\
                   & 2 & -32.69 & -48.07 & -57.69  \\
    \cmidrule(lr){1-5}
    1s-8x8   & 1 & -6.01 & -0.96 & -19.24  \\
                   & 2 & -7.42 & -0.79 & -19.34  \\
    \cmidrule(lr){1-5}
    1s-10x10 & 2 & -23.57 & -4.99 & -6.6  \\
                    & 3 & 2.32 & -6.28  & -7.48  \\
    \bottomrule
    \end{tabular}
    
    \label{tab:ben_lbf_targh}
\end{table}
\begin{table}[thbp]
    \centering
    \caption{Targeted Hadamard attack on RWARE and StarCraft. }
    \begin{tabular}{lc ccc}
    \toprule
    Task & $m$ & $\epsilon = 0.05$ & $\epsilon = 0.09$ & $\epsilon = 0.15$  \\
    \midrule
    tiny-4ag-sr-1   & 2 & 0.52 & -1.12  & -9.97  \\
                & 3 & -1.35 & -2.1  & -16.80  \\
    \cmidrule(lr){1-5}
    small-4ag-sr-1 & 2 & 3.26 & -11.61  & -12.70  \\
              & 3 & -2.17 & -14.51  & -19.23  \\
    \cmidrule(lr){1-5}
    10m\_vs\_11m    & 3 & -8.3 & -10.17 & -3.71  \\
    & 7 & -14.35 & 0.40  & 1.98  \\
    \cmidrule(lr){1-5}
    MMM2    & 3 & 3.26 & 0.73 & -6.32  \\
            & 7 & -2.76 & 1.22 & -7.10  \\
    \bottomrule
    \end{tabular}
    \label{tab:ben_rware_targh}
\end{table}

\subsection{Ablations}
\label{sec:ablation}

\subsubsection{Trained Agents} 
\begin{figure}[htbp]
    \centering
    \includegraphics[width=\columnwidth]{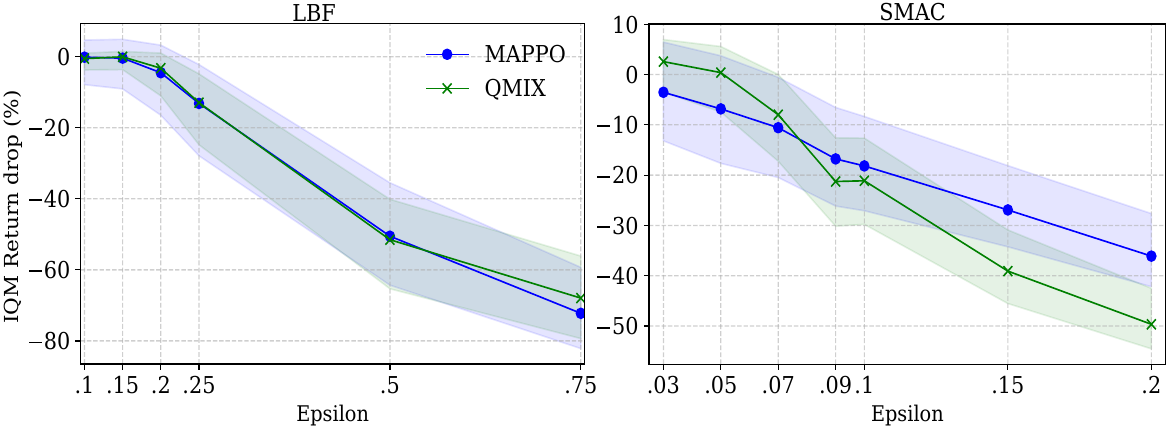}
    \caption{Performance across MARL algorithms.}
    \Description{Performance across MARL algorithms.}
    \label{fig:ben_lbf_algo}
\end{figure}

In Figure~\ref{fig:ben_lbf_algo}, we compare the performance of the Misalign attack against MAPPO and QMIX agents. The results indicate that our approach works for both policy-based and value-based algorithms. On the LBF benchmark, the performance is nearly identical. For StarCraft, however, the Misalign attack tends to be more impactful on QMIX agents at higher $\epsilon$ values. We do not report results for RWARE because QMIX did not converge.

\subsubsection{Data Collection} 
\begin{figure}[htbp]
    \centering
    \includegraphics[width=\columnwidth]{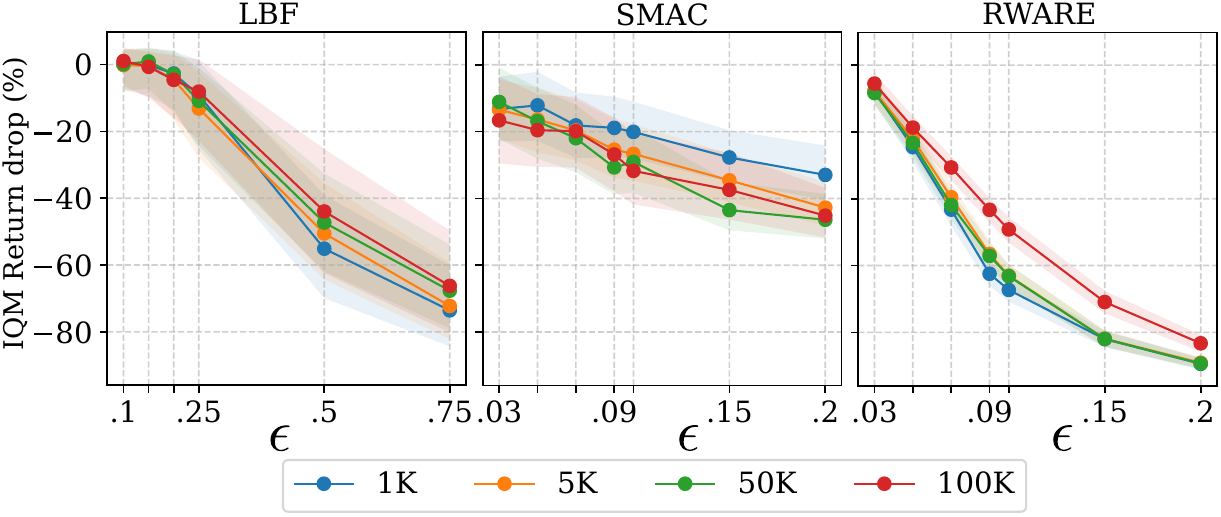}
    \caption{Effect of data collection size across benchmarks.}
    \Description{Collected data size effect across benchmarks.}
    \label{fig:ben_all_data}
\end{figure}

Figure~\ref{fig:ben_all_data} evaluates the Misalign attack with training data ranging from 1,000 to 100,000 steps. Surprisingly, strong attack performance is achieved with as few as 1,000 environment steps. Larger datasets do not consistently improve performance across the benchmarks, indicating that effective Misalign attacks can be generated with limited observation data. This reduces data collection requirements, which improves the stealth and the sample efficiency of the attack.

\subsubsection{Network Architecture}  
\begin{figure}[htbp]
    \centering
    \includegraphics[width=\columnwidth]{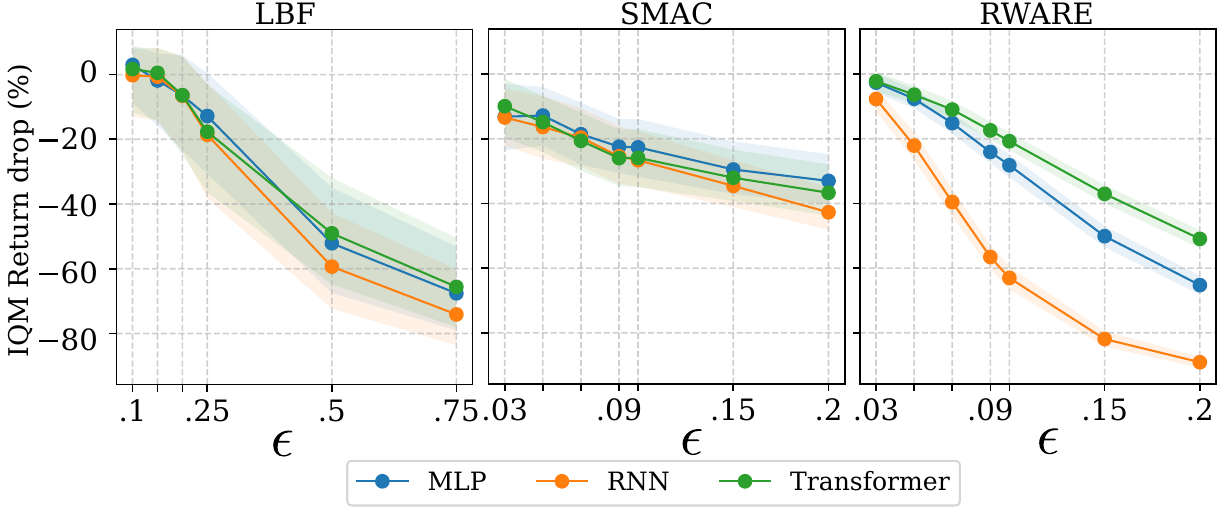}
    \caption{Effect of network architecture across benchmarks.}
    \Description{Neural architecture effect across benchmarks.}
    \label{fig:ben_all_net}
\end{figure}

Figure~\ref{fig:ben_all_net} compares different architectures for the alignment network $f_\theta$. RNNs generally achieve the strongest performance, with a particularly large margin on RWARE. MLPs and Transformers perform comparably. We tested deeper and more complex architectures, but they did not yield further performance gains.


\subsection{Discussion and Limitations} 

Most adversarial MARL work, including ours, evaluates attack effectiveness primarily using episodic return. We believe episode length is another important metric that is often overlooked, particularly in real-world scenarios with constrained adversaries.
In such settings, completely incapacitating the c-MARL system is not always realistic; a more plausible outcome is delaying task completion by increasing episode length, even if tasks are eventually completed. Longer episodes can substantially increase operational costs, including energy and computation, making this a meaningful measure of adversarial impact. We illustrate this using the Misalign attack on LBF. Table~\ref{tab:ben_lbf_episodelength} shows percentage increases in episode length, which often exceed double digits and reach up to 285\% in the worst case.

\begin{table}[htbp]
\centering
\caption{Percentage change in episode length on LBF tasks.}
\label{tab:ben_lbf_episodelength}
\setlength{\tabcolsep}{4pt}
\begin{tabular}{lrrrrrr}
\toprule
\textbf{Tasks} & \textbf{0.1} & \textbf{0.15} & \textbf{0.2} & \textbf{0.25} & \textbf{0.5} & \textbf{0.75} \\
\midrule
1s-8x8 & 1.30 & 1.55 & 2.09 & 3.63 & 12.09 & 12.18 \\
1s-10x10 & -5.68 & -9.66 & 2.26 & 7.94 & 16.64 & 22.88 \\
2s-8x8-coop & 9.59 & 43.55 & 138.52 & 152.49 & 170.86 & 170.86 \\
2s-8x8 & 5.17 & 18.84 & 31.93 & 44.26 & 72.24 & 72.24 \\
2s-10x10-coop & 34.44 & 23.78 & 65.78 & 68.52 & 79.04 & 85.19 \\
2s-10x10 & 12.45 & -2.72 & 14.36 & 20.78 & 52.38 & 54.29 \\
8x8-coop & 10.59 & 29.67 & 103.86 & 188.56 & 226.80 & 226.80 \\
8x8 & 5.94 & 21.06 & 51.93 & 75.93 & 193.21 & 285.80 \\
10x10-coop & 23.49 & 18.89 & 23.43 & 42.13 & 190.87 & 214.86 \\
10x10 & 0.66 & 20.32 & 11.93 & 29.41 & 49.43 & 110.93 \\
\bottomrule
\end{tabular}
\end{table}
The effectiveness of the Misalign attack is influenced by several factors commonly encountered in MARL settings, including partial observability, which we discuss above. In addition, issues like handling heterogeneous agents and multi-modal observations may further affect the effectiveness of our attack. Moreover, we studied attacks from the perspective of the adversary’s knowledge. However, other perspectives can also be considered, including system-level constraints, detectability in real-world settings, and potential defenses. We leave these directions for future work.

\section{Conclusion}
In this paper, we empirically investigate whether effective black-box attacks against collaborative MARL can be generated without the information required to train a surrogate policy. We present two adversarial attacks under highly constrained attacker assumptions. In the first threat model, the adversary can collect and perturb the observations. Our approach does not rely on training surrogate policies or exploiting adversarial transferability. The second threat model considers an even more constrained scenario in which the adversary can only perturb but not collect the observations. Our attacks are motivated by the reliance of collaborative agents on aligned perceptions for effective cooperation and aim to disrupt this alignment through small observation perturbations. Our empirical evaluation across 22 tasks demonstrates that effective attacks remain possible under both threat models. In several tasks, our attacks match or outperform the imitation attack despite requiring less information. Notably, our method requires minimal data to generate effective perturbations. We found that as few as 1,000 steps are sufficient to generate effective perturbations.






\bibliographystyle{ACM-Reference-Format} 
\bibliography{sample}


\clearpage
\onecolumn
\begin{center}
  {\LARGE\bfseries Technical Appendix\par}
\end{center}

\appendix

\section{Experimental Details}

We provide the full details on selected environments in Tables \ref{tab:selected_lbf_envs}, \ref{tab:rwr_envs}, and \ref{tab:smac_envs}. We provide details on observability, dimensionality, and the number of agents.

We provide the used hyperparameters in tables \ref{tab:qmic_hyper}, \ref{tab:mappo_smac_hyper}, \ref{tab:mappo_lbf_hyper}
, and \ref{tab:mappo_rware_hyper}.  We use the Epymarl library for training. We only report hyperparameters that are different from the default values found in Epymarl configurations. For the RWARE benchmark, we only train MAPPO agents, as QMIX  performs poorly at this task. Once the training is complete, we freeze the policy weights.

For the Misalign attack, the adversary starts by collecting a dataset of observations from 5,000 environment transitions (i.e., $\mathcal{T}^c =5000$). This dataset is subsequently used to train the neural network $f_\theta$. We test with three different architectures:

\begin{itemize}
    \item Feedforward neural network: It consists of three hidden layers with 1024 hidden units.
    \item Recurrent neural network: For LBF and StarCraft, we use one recurrent layer with a hidden dimension of 1024. For RWARE, we use 2 consecutive layers.
    \item  Encoder-only transformer: For RWARE and LBF, we use d\_model=256 and dim\_feedforward=2048 with 16 heads. For StarCraft, we use d\_model=128 and dim\_feedforward=1024 with 8 heads.
\end{itemize}

We use a learning rate of $0.0001$ and a batch size of 64. The feedforward and transformer networks are trained for 100 epochs, while the RNN network is trained for 300 epochs.

\begin{table}[htbp]
    
  \centering
  \begin{tabular}{lccccc}
  \hline
  Environment &  FO & PO &  HC &  $\text{dim}(o_i)$ & N\\\hline
  8x8-3p-2f & \ding{51} & \ding{55} &\ding{55} & 15 & 3\\
  8x8-3p-2f-coop & \ding{51} & \ding{55} &\ding{51} & 15 & 3\\
  1s-8x8-3p-2f & \ding{55} & \ding{51} &\ding{55} & 15 & 3\\
  2s-8x8-3p-2f & \ding{55} & \ding{51} &\ding{55} & 15 & 3\\
  2s-8x8-3p-2f-coop & \ding{55} & \ding{51} &\ding{51} & 15 & 3\\
  10x10-4p-2f & \ding{51} & \ding{55} &\ding{55} &  18 & 4\\
  10x10-4p-2f-coop & \ding{51} & \ding{55} &\ding{51} & 18& 4 \\
  1s-10x10-4p-2f & \ding{55} & \ding{51} &\ding{55} &  18 & 4\\
  2s-10x10-4p-2f & \ding{55} & \ding{51} &\ding{55} & 18 & 4\\
  2s-10x10-4p-2f-coop & \ding{55} & \ding{51} &\ding{51} & 18 & 4 \\
  \hline
  \end{tabular}
  \caption{Selected LBF environments: we select (FO) fully observable, (PO) partially observable, and (HC) highly cooperative tasks.}
  \label{tab:lbf_envs}
  \label{tab:selected_lbf_envs}
\end{table}

\begin{table}[htbp]
  \centering
  \begin{tabular}{lcccc}
  \hline
  Environment &  PO & SR &  $\text{dim}(o_i)$ & N \\\hline
  tiny-4ag & \ding{51} & $3\times3$ & 71 & 4\\
  tiny-4ag-sr2 & \ding{51} & $5\times5$ & 183 & 4\\
  tiny-4ag-sr3 & \ding{51} & $7\times7$ & 351 & 4\\
  small-4ag & \ding{51} & $3\times3$ & 71 & 4 \\
  small-4ag-sr2& \ding{51} & $5\times5$ & 183 & 4\\
  small-4ag-sr3 & \ding{51}  & $7\times7$ &  351 & 4\\
  \hline
  \end{tabular}
  \caption{Selected RWARE environments: we select sensory range (SR) of $3\times3$, $5\times5$ and $7\times7$. }
  \label{tab:rwr_envs}
\end{table}

\begin{table}[htbp]
  \centering
  \begin{tabular}{lccc}
    \hline
    Environment &  PO  &  $\text{dim}(o_i)$ & N \\\hline
    27m\_vs\_30m & \ding{51} &  285 & 27\\
    10m\_vs\_11m & \ding{51} &  105 & 10\\
    MMM2 & \ding{51} &  176 & 10\\
    protoss\_5\_vs\_5 & \ding{51}  & 92 & 5 \\
    terran\_5\_vs\_5 & \ding{51}  & 82 & 5\\
    zerg\_5\_vs\_5 & \ding{51}   &  82 & 5\\
    \hline
  \end{tabular}
  \caption{Selected StarCraft environments: all environments are partially observable (PO)}
  \label{tab:smac_envs}
\end{table}

\begin{table}[htbp]
    \centering
    \begin{tabular}{llll}
    \hline
    \textbf{Hyperparameter} & \textbf{SMAC} & \textbf{LBF} \\
    \hline
    Runner type & parallel  & episode \\
    Batch size per run & 8  & 1 \\
    Network hidden dimension & 128 & 64\\
    Learning rate & 0.001 & 0.0003 \\
    Standardise rewards & True & True \\
    Use RNN & True  & True \\
    Epsilon anneal time & 50000  & 200000 \\
    Target update interval & 200 & 0.01 \\
    Training batch size & 64 & 32 \\
    Maximum timesteps ($t_{max}$) & 20,050,000 & 5050000 \\
    \hline
    \end{tabular}
    \caption{QMIX Hyperparameters Used for StarCraft and LBF benchmark}
    \label{tab:qmic_hyper}
    \end{table}

\begin{table}[htbp]
    
    \centering
    \begingroup
    \setlength{\tabcolsep}{3pt} 
    \begin{tabular}{lccccccc}
    \hline
    \textbf{Environments} & hdim$^{1}$ & lr & Stand$^{2}$ & ent & q\_nstep  & Buffer & Epochs \\
    \hline
    
    MMM2 & 64 & 0.0005 & False & 0.001 & 10  & 10 & 4 \\
    27m\_vs\_30m & 64 & 0.0005 & False & 0.001 & 10  & 10 & 4 \\
    10m\_vs\_11m & 128 & 0.001 & True & 0 & 5 & 8 & 10 \\
    protoss\_5\_vs\_5  & 64 & 0.0005 & False & 0.001 & 10   & 10 & 4  \\
    terran\_5\_vs\_5 & 128 & 0.0003  & True & 0.001 & 5  & 10 & 4  \\
    zerg\_5\_vs\_5 & 128 & 0.0003  & True & 0.001 & 5  & 10 & 4  \\
    \hline
    \end{tabular}
    \endgroup
    \caption{MAPPO Hyperparameters for StarCraft benchmark.}
    \label{tab:mappo_smac_hyper}
\end{table}

\begin{table}[htbp]
    
    \centering
    \begin{tabular}{lccccc}
        \hline
        Environment         &  lr  & hdim &  Stand &  RNN  & q\_nstep\\\hline
        8x8-3p-2f           & 0.0003 & 128 & False & False & 5\\
        8x8-3p-2f-coop        & 0.0005 & 32 &True& True & 5\\
        1s-8x8-3p-2f       & 0.0003 & 128 &False & True & 10\\
        2s-8x8-3p-2f         & 0.0003 & 128  &False & False & 5\\
        2s-8x8-3p-2f-coop   & 0.0005 & 128 &True & True & 5\\
        10x10-4p-2f         & 0.0003 & 128 &False &  False & 5\\
        10x10-4p-2f-coop       & 0.0003 & 32 &True & True & 5\\
        1s-10x10-4p-2f    & 0.0003 & 128 &False & True& 10 \\
        2s-10x10-4p-2f       & 0.0003 & 128 &False &  False & 5\\
        2s-10x10-4p-2f-coop & 0.0005 & 128 &True & False & 5 \\
        \hline
    \end{tabular}
    \caption{MAPPO Hyperparameters for LBF benchmark. }
    \label{tab:mappo_lbf_hyper}
\end{table}

\begin{table}[htbp]
    \centering
    \begin{tabular}{lccccc}
    \hline
        Environment &  lr & Stand &  RNN & q\_nstep & $t_{max}$ \\\hline
        tiny-4ag & 0.0005 & False & False & 10 & 25M\\
        tiny-4ag-sr2 & 0.0008 & False & False & 10 &25M\\
        tiny-4ag-sr3 & 0.0008 & False & False & 10 &25M\\
        small-4ag & 0.0005 & False & False & 10 &25M \\
        small-4ag-sr2& 0.0008 & False & True & 10 &25M\\
        small-4ag-sr3 & 0.0008  & False &  False & 20 &40M\\
    \hline
  \end{tabular}
  \caption{MAPPO Hyperparameters for RWARE. }
  \label{tab:mappo_rware_hyper}
\end{table}

We test the following values of attack budget:
\[
\begin{aligned}
\epsilon &\in \{0.03,0.05,0.07,0.09,0.1,0.15,0.2\} &&\text{for RWARE and StarCraft},\\
\epsilon &\in \{0.1,0.15,0.2,0.25,0.5,0.75\} &&\text{for LBF}.
\end{aligned}
\]

\subsection{Additional results}

\begin{figure*}[htbp]
    \centering
    \includegraphics[width=\textwidth]{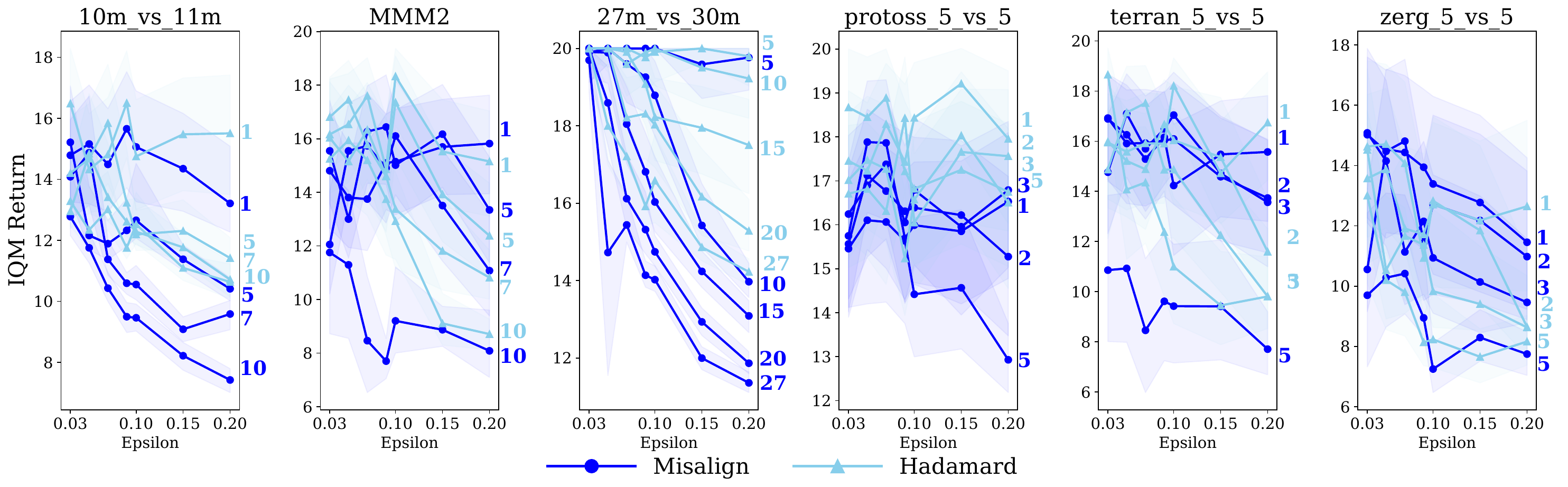}
    \caption{Performance of targeted Misalign attacks on StarCraft.}
    \Description{Performance of targeted Misalign attacks on StarCraft.}
    \label{fig:ben_smac_light_all}
\end{figure*}

\begin{figure*}[htbp]
    \centering
    \includegraphics[width=\textwidth]{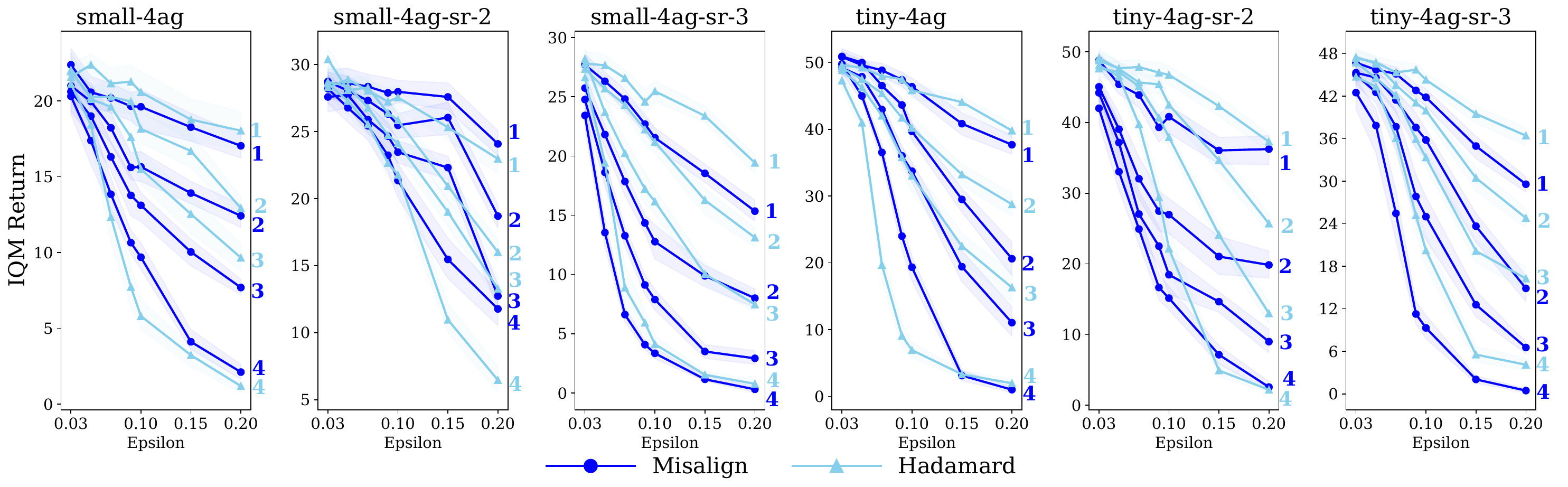}
    \caption{Performance of targeted Misalign attacks on RWARE. }
    \Description{Performance of targeted Misalign attacks on RWARE.}
    \label{fig:ben_rware_light_all}
\end{figure*}

\begin{figure*}[htbp]
    \centering
    \includegraphics[width=\textwidth]{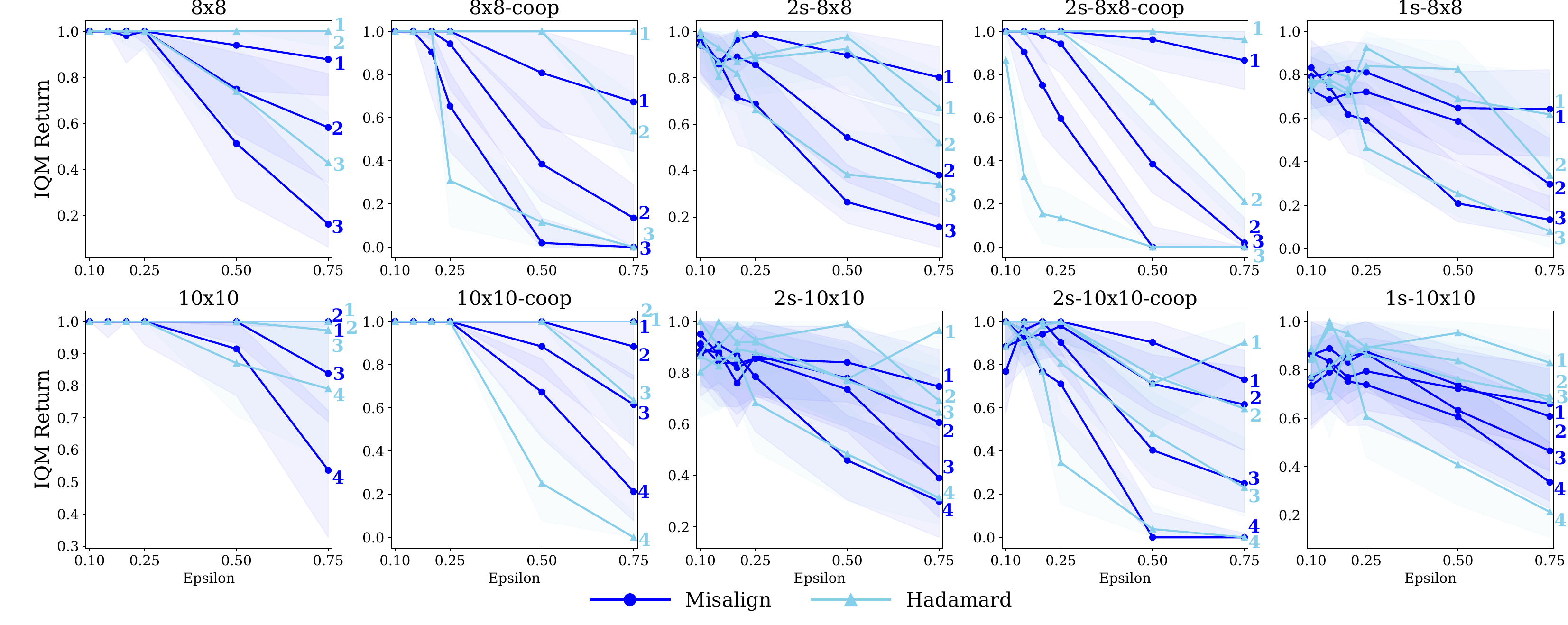}
    \caption{Performance of targeted Misalign attacks on LBF.}
    \Description{Performance of targeted Misalign attacks on LBF.}
    \label{fig:ben_lbf_light_all}
\end{figure*}

We report the full results of targeted attacks from Section 4.3. Figures \ref{fig:ben_smac_light_all}, \ref{fig:ben_rware_light_all}, and \ref{fig:ben_lbf_light_all} report the full results of targeted Misalign attack. In Tables \ref{tab:ben_lbf_targh_all}, \ref{tab:ben_rware_targh_all}, and \ref{tab:ben_smac_targh_all} we report the full results of combining Hadamard and Misalign attack.


\begin{table}[htbp]
    \centering
    \begin{tabular}{lc ccc}
    \toprule
    Task & $m$ & $\epsilon = 0.15$ & $\epsilon = 0.2$ & $\epsilon = 0.25$  \\
    \cmidrule(lr){1-5}
    1s-10x10-4p-2f & 2 & -23.57 & -4.99 & -6.6  \\
                   & 3 & 2.32 & -6.28  & -7.48  \\
    \cmidrule(lr){1-5}
    2s-10x10-4p-2f & 2 & -10.02 & -21 & -8,57  \\
                   & 3 & -10.49 & -11.71  & -15.26  \\
    \cmidrule(lr){1-5}
    2s-10x10-4p-2f-coop & 2 & -3.84 & -15.36 & -17.30  \\
                   & 3 & -3.84 & -13.46  & -15.38  \\
    \cmidrule(lr){1-5}
    1s-8x8-3p-2f   & 1 & -6.01 & -0.96 & -19.24  \\
                   & 2 & -7.42 & -0.79 & -19.34  \\
    \cmidrule(lr){1-5}
    2s-8x8-3p-2f   & 1 & -0.93 & 3.15 & 1.88  \\
                   & 2 & 3.75 & -18.10 & -11.17  \\
    \cmidrule(lr){1-5}
    2s-8x8-3p-2f-coop   & 1 & 0 & -3.84 & -3.84  \\
                   & 2 & -32.69 & -48.07 & -57.69  \\
    \bottomrule
    \end{tabular}
    \caption{ Targeted Hadamard attack on LBF. }
    \label{tab:ben_lbf_targh_all}
\end{table}

\begin{table}[htbp]
    \centering
    \begin{tabular}{lc ccc}
    \toprule
    Task & $m$ & $\epsilon = 0.05$ & $\epsilon = 0.09$ & $\epsilon = 0.15$  \\
    \cmidrule(lr){1-5}
    tiny-4ag    & 1 & 0.22 & -0.37 & -4.50  \\
                & 2 & 0.52 & -1.12  & -9.97  \\
                & 3 & -1.35 & -2.1  & -16.80  \\
    \cmidrule(lr){1-5}
    small-4ag & 1 & -9.80 & -6.71 & -2.35  \\
              & 2 & 3.26 & -11.61  & -12.70  \\
              & 3 & -2.17 & -14.51  & -19.23  \\
    \bottomrule
    \end{tabular}
    \caption{Targeted Hadamard attack on RWARE. }
    \label{tab:ben_rware_targh_all}
\end{table}

\begin{table}[htbp]
    \centering
    \begin{tabular}{lc ccc}
    \toprule
    Task & $m$ & $\epsilon = 0.05$ & $\epsilon = 0.09$ & $\epsilon = 0.15$  \\
    \cmidrule(lr){1-5}
    10m\_vs\_11m    & 3 & -8.3 & -10.17 & -3.71  \\
                    & 7 & -14.35 & 0.40  & 1.98  \\
    \cmidrule(lr){1-5}
    MMM2            & 3 & 3.26 & 0.73 & -6.32  \\
                    & 7 & -2.76 & 1.22 & -7.10  \\
    \cmidrule(lr){1-5}
    protoss\_5\_vs\_5 & 2 & -4.40 & -7.58 & -4.10  \\
                      & 3 & -10.10 & -1.37 & -4.88  \\
    \cmidrule(lr){1-5}
    terran\_5\_vs\_5 & 2 & 4.05 & -3.05 & -19.48  \\
                      & 3 & -12.35 & -21.03 & -5.15  \\
    \cmidrule(lr){1-5}
    zerg\_5\_vs\_5 & 2 & -22.28 & -11.09 & -3.77  \\
                      & 3 & 0.27 & -12.86 & -4.40  \\
    \bottomrule
    \end{tabular}
    \caption{Targeted Hadamard attack on StarCraft. }
    \label{tab:ben_smac_targh_all}
\end{table}

\end{document}